%% file: main.tex
\documentclass[10pt,twocolumn,letterpaper]{article}

\usepackage[pagenumbers]{cvpr} 


\definecolor{cvprblue}{rgb}{0.21,0.49,0.74}
\usepackage[pagebackref,breaklinks,colorlinks,allcolors=cvprblue]{hyperref}
\usepackage[linesnumbered,ruled]{algorithm2e}
\usepackage{amsmath}

\DeclareMathOperator*{\argmax}{arg\,max}
\newcommand{\tsc}[1]{\textsuperscript{#1}}

\def\paperID{*****} 
\def\confName{CVPR}
\def\confYear{2025}

\title{Extending SEEDS to a Supervoxel Algorithm for Medical Image Analysis}

\author{Chenhui Zhao\tsc{1}\quad
Yan Jiang\tsc{2}\quad
Todd C. Hollon\tsc{1}\\[1em]
\tsc{1}University of Michigan\quad
\tsc{2}Beijing University of Posts and Telecommunications
}

\begin{document}
\maketitle
\begin{abstract}
    In this work, we extend the SEEDS~\cite{van2015seeds} superpixel algorithm from 2D images to 3D volumes, resulting in 3D SEEDS, a faster, better, and open-source supervoxel algorithm for medical image analysis. 
    We compare 3D SEEDS with the widely used supervoxel algorithm SLIC~\cite{6205760} on 13 segmentation tasks across 10 organs.
    3D SEEDS accelerates supervoxel generation by a factor of 10, improves the achievable Dice score by +6.5\%, and reduces the under-segmentation error by -0.16\%. 
    The code is available at \url{https://github.com/Zch0414/3d_seeds}
\end{abstract}

\section{Introduction}
\label{sec: introduction}

In medical image analysis, supervoxel algorithms aim to over-segment volumetric data by grouping voxels that share a predefined property, \textit{e.g.}, intensity homogeneity.
Numerous supervoxel algorithms~\cite{8392300, ouyang2020self, 6205760, liu2016manifold, amami2019adaslic} have been proposed, typically based on either gradient-ascent~\cite{6205760, liu2016manifold, amami2019adaslic} or graph-cut~\cite{8392300, ouyang2020self} approaches.
To date, The SLIC superpixel algorithm~\cite{6205760} clusters pixels based on color similarity and spatial proximity, efficiently generating compact and uniform superpixels. It can also be seamlessly extended to a supervoxel algorithm~\cite{tamajka2016automatic} for volumetric data. Several methods based on SLIC have been proposed, either originally aiming to generate superpixels~\cite{liu2016manifold} or supervoxels~\cite{amami2019adaslic}. ADNet~\cite{ouyang2020self} extends the Felzenszwalb superpixel algorithm~\cite{felzenszwalb2004efficient} to 3D supervoxels, enabling greater diversity in shape. Another graph-based approach~\cite{8392300} formulates supervoxel segmentation as a graph partitioning problem, aiming to optimize a specific objective function for efficient and effective segmentation.

Nevertheless, in recent deep-learning-based medical image analysis approaches~\cite{tamajka2016automatic, tian2017supervoxel, 7986891, ma2020boundary} that utilize supervoxels, SLIC~\cite{6205760} remains the most widely used supervoxel algorithm for two key reasons:
1. SLIC is still the state-of-the-art algorithm in many over-segmentation tasks~\cite{8392300}.
2. SLIC is open-source in Python, the primary programming language used in most deep learning frameworks.
However, supervoxels can only be generated offline with the SLIC algorithm due to its speed, limiting the use of online data augmentation methods in deep learning frameworks, such as random cropping, rotation, or intensity scaling and shifting.
Moreover, while many studies~\cite{tamajka2016automatic, tian2017supervoxel, 7986891, ma2020boundary} have highlighted the advantages of using SLIC over raw voxels, some works~\cite{ouyang2020self, ke2023learning} reports that SLIC may not be the optimal algorithm for large-scale self-supervised pre-training.

\begin{figure}[!t]
\centering
    \includegraphics[width=0.96\columnwidth]{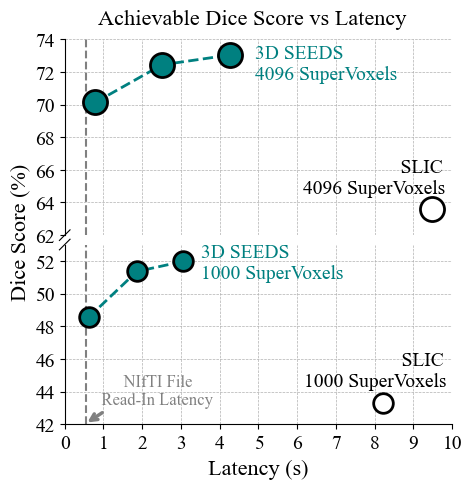}
    \caption{\textbf{Illustration of the over-segmentation performance.} Compared to the SLIC algorithm, 3D SEEDS accelerates supervoxel generation and enhances the achievable Dice score on 13 segmentation tasks across 10 organs.}
    \label{fig: intro}
\end{figure}

In this work, we address the unmet need for online supervoxel generation, as posed by deep learning frameworks.
Compared to supervoxel algorithms, superpixel algorithms~\cite{ren2003learning, felzenszwalb2004efficient, 6205760, van2015seeds} have been more extensively studied by researchers. Among these, SEEDS~\cite{van2015seeds} stands out as a recent and promising superpixel algorithm, achieving near real-time generation on a single CPU.
To achieve both high performance and speed, we extend SEEDS~\cite{van2015seeds} to a supervoxel algorithm, namely \textbf{3D SEEDS}.

While adhering as closely as possible to the original algorithm, extending SEEDS to volumetric data remains non-trivial due to the inherent nature of SEEDS.
The algorithm adjusts the boundary of each superpixel by transferring boundary pixels to neighboring superpixels, which can potentially split a superpixel into two separate parts.
Avoiding this can be challenging, especially for supervoxels in 3D space, where the geometry is significantly more complex than that of superpixels in 2D space. 
To this end, we identify 16 cases that must be avoided when implementing the 3D SEEDS algorithm. 3D SEEDS is implemented based on the SEEDS algorithm provided by OpenCV\footnote[1]{\url{https://opencv.org/}} and is compatible with both C++ and Python programming languages.

We evaluate 3D SEEDS on the BraTS~\cite{baid2021rsna} and BTCV\footnote[2]{\url{https://www.synapse.org/Synapse:syn3193805/}} benchmarks, involving 13 segmentation tasks across 10 organs.
To demonstrate the over-segmentation performance of 3D SEEDS in the context of recent progress, we introduce a new metric, the achievable Dice score, alongside the previously proposed metric, under-segmentation error~\cite{van2015seeds}.
As shown in Figure~\ref{fig: intro}, when compared with the widely used SLIC supervoxel algorithm~\cite{6205760}, 3D SEEDS can:
\begin{itemize}
    \item Enhance the achievable Dice score on 13 segmentation tasks across 10 organs.
    \item Achieve higher over-segmentation performance by increasing the number of supervoxels.
    \item Accelerate the supervoxel generation to a speed comparable to the NIfTI file read-in speed.
\end{itemize}

\section{SEEDS}
\label{sec: seeds}
In this section, we revisit the SEEDS superpixel algorithm~\cite{van2015seeds} from an implementation perspective, closely adhering to the implementation provided by the OpenCV\footnotemark[1].

\subsection{Overview}
\label{subsec: seeds overview}

SEEDS recursively extracts superpixels $\{A_{k}\}^{K}_{k=1}$ from a regular grid. SEEDS introduces two update strategies: pixel-level and block-level updates. For details on these strategies, we refer readers to the original paper~\cite{van2015seeds} and additional resources\footnote[3]{\url{https://davidstutz.de/}}.
We outline one iteration in Algorithm~\ref{alg: seeds}. Given a superpixel $A_{k}$, $A^{l}_{k}$ denotes either a single pixel~(pixel-level update) or a small set of pixels~(block-level update) located at the boundary of $A_{k}$. 
SEEDS adjusts the boundary of $A_{k}$ by transferring $A^{l}_{k}$ to a neighboring superpixel $A_{n}$. 
Two conditions govern the transfer: A 2D check-splitting function $\mathcal{S}_{2D}(\cdot, \cdot)$ and an energy function $\mathcal{E}(\cdot, \cdot)$.

\begin{algorithm}[!t]
\caption{One Iteration in SEEDS~\cite{van2015seeds}}\label{alg: seeds}
\SetKw{Propose}{Propose}
\SetKw{and}{and}
\SetKw{Horizontal}{Horizontal}
\SetKw{Vertical}{Vertical}
\KwIn{Original $\{A_{k}\}^{K}_{k=1}$}
\KwOut{Updated $\{A_{k}\}^{K}_{k=1}$}
\For{$A_{k}$ in $\{A_{k}\}^{K}_{k=1}$}{
    \Propose{$A_{n} \in \{\Horizontal, \Vertical\}$}\;
    \Propose{$A^{l}_{k}$}\;
    \If{$\mathcal{S}_{2D}(A_{k}, A^{l}_{k})$ \and $\mathcal{E}(A_{k} \setminus A^{l}_{k}, A^{l}_{k}) < \mathcal{E}(A_{n}, A^{l}_{k})$}{
        $A_{k} \gets A_{k} \setminus A^{l}_{k}$\; 
        $A_{n} \gets A_{n} \cup A^{l}_{k}$\;
    }
}
\Return{$\{A_{k}\}_{k=1}^{K}$}
\end{algorithm}

\begin{figure}[!t]
\centering
    \includegraphics[width=0.8\columnwidth]{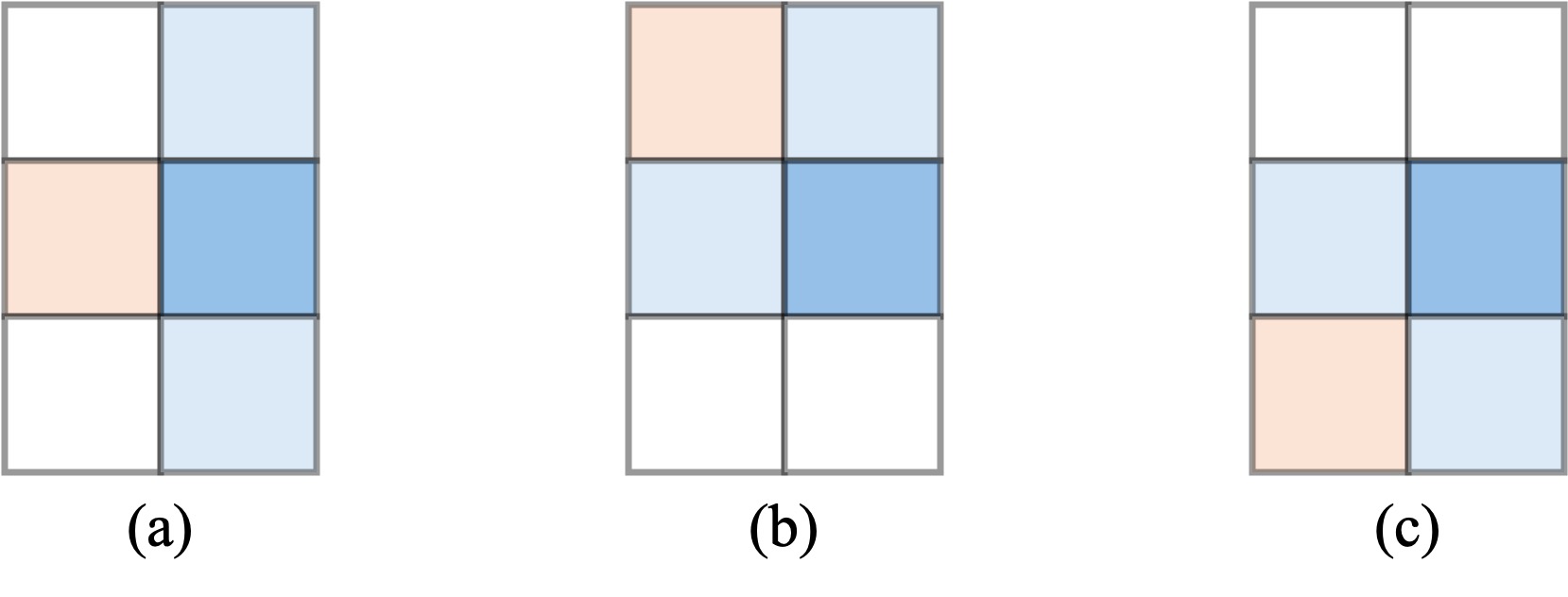}
    \caption{\textbf{Illustration of the 2D check-split function.} The deeper blue square represents the portion~($A^{l}_{k}$) of a superpixel~($A_{k}$) to be transferred to another superpixel~($A_{n}$) in the horizontal direction~(from left to right). The lighter blue square indicates the remaining part of $A_{k}$, while the orange square belongs to any other superpixel except for $A_{k}$. The 2D check-split function returns false if cases (a), (b), or (c) occur; otherwise, it returns true.}
    \label{fig: checksplit2d}
\end{figure}

\begin{figure}[!t]
\centering
    \includegraphics[width=0.8\columnwidth]{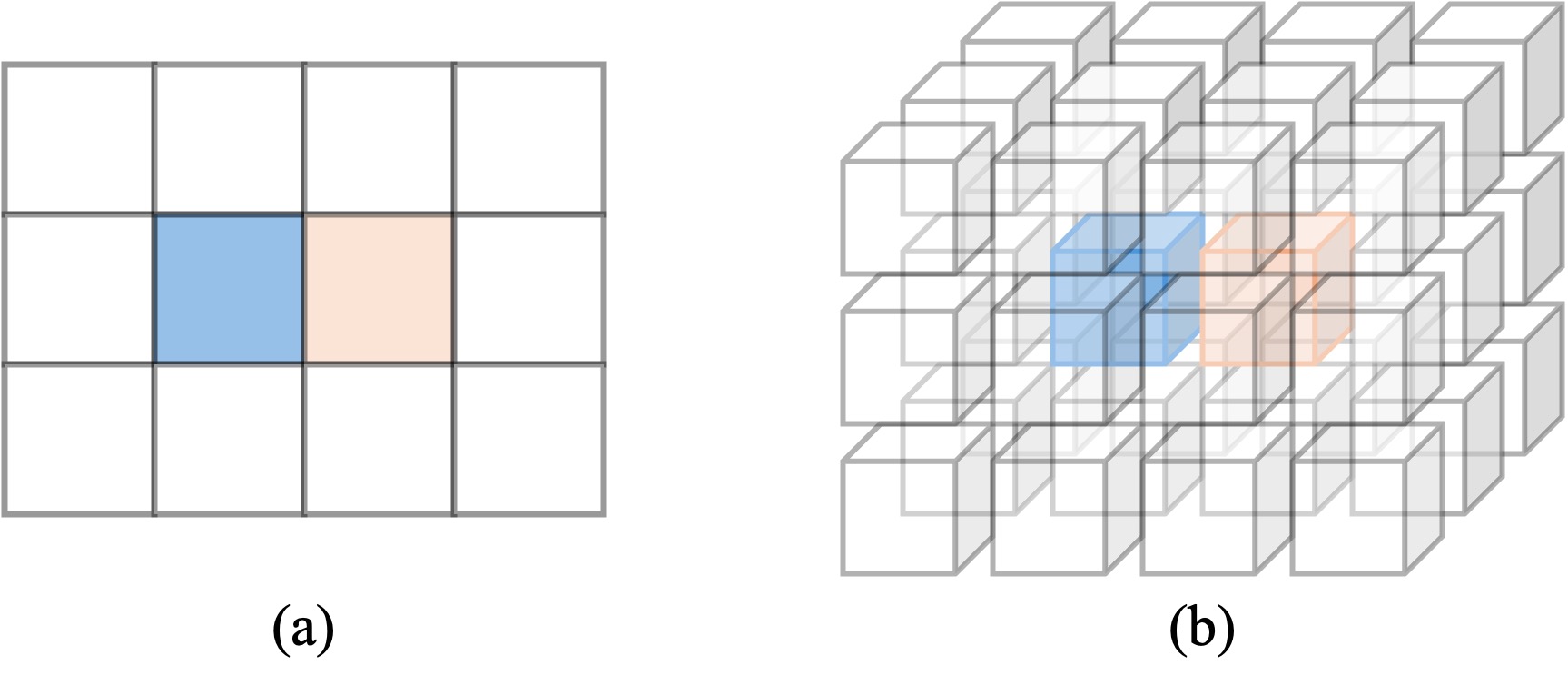}
    \caption{\textbf{Illustration of $N_{\mathcal{A}^{l}_{k}}$ in the boundary term.} (a) The deeper blue square represents a single pixel $A^{l}_{k}$ in the superpxiel $A_{k}$, which is to be transferred to another superpixel in the horizontal direction~(from left to right). $N_{\mathcal{A}^{l}_{k}}$ is computed within the $3 \times 4$ area surrounding $A^{l}_{k}$, ignoring the orange square. (b) $N_{\mathcal{A}^{l}_{k}}$ is computed in the 3D space.}
    \label{fig: boundary term}
\end{figure}

\subsection{2D Check-Split Function}
\label{subsec: checksplit2d}

The 2D check-split function $\mathcal{S}_{2D}(\cdot, \cdot)$ ensures that any transfer of $A^{l}_{k}$ does not split the superpixel $A_{k}$ into two separate parts. 
When transferring $A^{l}_{k}$ along the horizontal direction (from left to right), Figure~\ref{fig: checksplit2d} showcases 3 scenarios where $\mathcal{S}_{2D}(A_{k}, A^{l}_{k})$ returns false; otherwise, it returns true.
For other directions, the corresponding illustrations can be obtained by rotating scenarios (a), (b), and (c) from Figure~\ref{fig: checksplit2d} in the 2D space.

\begin{figure*}[!t]
\centering
    \includegraphics[width=0.9\textwidth]{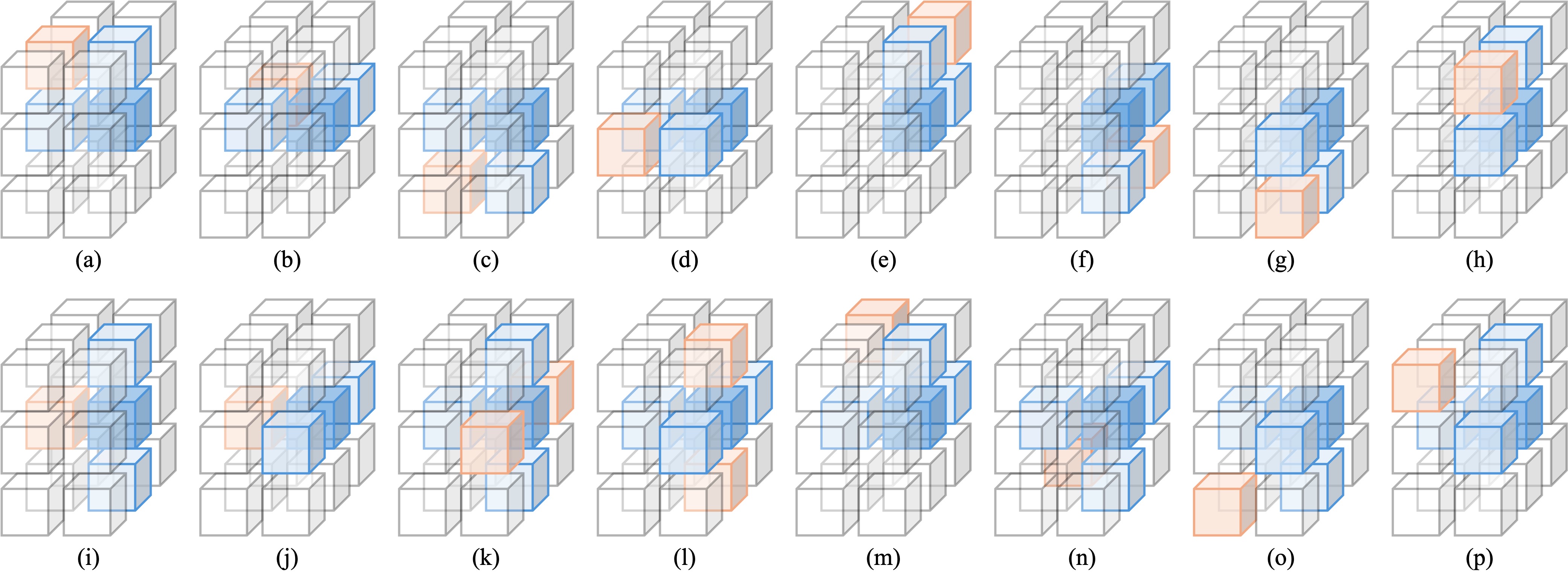}
    \caption{\textbf{Illustration of the 3D check-split function.} The deeper blue cube represents the portion~($A^{l}_{k}$) of a supervoxel~($A_{k}$) to be transferred to another supervoxel~($A_{n}$) in the sagittal direction~(from left to right). The lighter blue cube represents the remaining part of $A_{k}$, while the orange cube belongs to any other supervoxel except for $A_{k}$. The 3D check-split function returns false if cases (a)-(p) occur; otherwise, it returns true.}
    \label{fig: checksplit3d}
\end{figure*}

\subsection{Energy Function}
\label{subsec: energy function}

The energy function $\mathcal{E}(\cdot, \cdot)$ is defined as the product of two terms: 1. The color distribution term $\mathcal{H}(\cdot, \cdot)$, which is based on the color likelihood of the superpixel. 2. The boundary term $\mathcal{G}(\cdot, \cdot)$, which is based on the boundary smoothness of the superpixel. 
Given two sets of pixels $A^{a}_{k}$ and $A^{b}_{n}$, the energy is computed as: 
\begin{equation}
    \mathcal{E}(A^{a}_{k}, A^{b}_{n}) = \mathcal{H}(A^{a}_{k}, A^{b}_{n}) \times \mathcal{G}^{\lambda}(A^{a}_{k}, A^{b}_{n})
\label{equ: energy function}
\end{equation}
where $\lambda$ weights the influence of the boundary term.

\noindent\textbf{Color Distribution Term.}
Given a set of pixels $A^{l}_{k}$, $C_{\mathcal{A}^{l}_{k}}(j)$ denotes the number of pixels located in the $j$-th color bin. The color distribution term between $A^{a}_{k}$ and $A^{b}_{n}$ is computed as: 
\begin{equation}
    \mathcal{H}(A^{a}_{k}, A^{b}_{n}) = \sum_{j}{\min(C_{\mathcal{A}^{a}_{k}}(j), C_{\mathcal{A}^{b}_{n}}(j))}
\label{equ: color distribution term}
\end{equation}

\noindent\textbf{Boundary Term.}
The boundary term is computed only for the individual pixel. Given a single pixel $A^{l}_{k}$, as shown in Figure~\ref{fig: boundary term}~(a), $N_{\mathcal{A}^{l}_{k}}$ denotes the number of pixels in $A_{k}$ surrounding $A^{l}_{k}$. Note that $A^{l}_{k}$ is located in the $A^{l}_{k}$-th color bin. The boundary term between two individual pixels $A^{a}_{k}$ and $A^{b}_{n}$ is computed as:
\begin{equation}
    \mathcal{G}(A^{a}_{k}, A^{b}_{n}) = N_{\mathcal{A}^{b}_{n}} \times C_{\mathcal{A}^{b}_{n}}(A^{a}_{k})
\label{equ: boundary term}
\end{equation}

\section{3D SEEDS}
\label{sec: seeds3d}
In this section, we present the 3D SEEDS supervoxel algorithm by extending SEEDS from 2D images to 3D volumes.

\subsection{Overview}
\label{subsec: seeds3d overview}

We closely adhere to the original SEEDS implementation. In Algorithm~\ref{alg: seeds3d}, we highlight the modification in {\color{blue}{blue}}: 1. Given a supervoxel $A_{k}$, the neighboring supervoxel $A_{n}$ is proposed in three directions: sagittal, coronal, and axial. 2. The 3D check-split function $\mathcal{S}_{3D}(\cdot, \cdot)$. All other parts of the algorithm remain unchanged, including the energy function and the update strategy. However, note that as shown in Figure~\ref{fig: boundary term}~(b), $N_{\mathcal{A}^{l}_{k}}$ is computed in the 3D space for the boundary term in the energy function.

\begin{algorithm}[!t]
\caption{One Iteration in \color{blue}{3D SEEDS}}\label{alg: seeds3d}
\SetKw{Propose}{Propose}
\SetKw{and}{and}
\SetKw{Sagittal}{Sagittal}
\SetKw{Coronal}{Coronal}
\SetKw{Axial}{Axial}
\KwIn{Original $\{A_{k}\}^{K}_{k=1}$}
\KwOut{Updated $\{A_{k}\}^{K}_{k=1}$}
\For{$A_{k}$ in $\{A_{k}\}^{K}_{k=1}$}{
    \Propose{$A_{n} \in \{\color{blue}{\Sagittal, \Coronal, \Axial}\}$}\;
    \Propose{$A^{l}_{k}$}\;
    \If{$\color{blue}{\mathcal{S}_{3D}(A_{k}, A^{l}_{k})}$ \and $\mathcal{E}(A_{k} \setminus A^{l}_{k}, A^{l}_{k}) < \mathcal{E}(A_{n}, A^{l}_{k})$}{
        $A_{k} \gets A_{k} \setminus A^{l}_{k}$\; 
        $A_{n} \gets A_{n} \cup A^{l}_{k}$\;
    }
}
\Return{$\{A_{k}\}_{k=1}^{K}$}
\end{algorithm}

\subsection{3D Check-Splitting Function}
\label{subsec: checksplit3d}

The 3D check-split function $\mathcal{S}_{3D}(\cdot, \cdot)$ ensures that any transfer of $A^{l}_{k}$ does not split the supervoxel $A_{k}$ into two separate parts in the 3D space. 
The geometry of a supervoxel is significantly more complex than that of a superpixel. However, we find that the scenarios illustrated in Figure~\ref{fig: checksplit3d} are sufficient to prevent any potential splits.
When transferring $A^{l}_{k}$ along the sagittal direction~(from left to right), $\mathcal{S}_{3D}(A_{k}, A^{l}_{k})$ returns false if cases (a)-(p) in Figure~\ref{fig: checksplit3d} occur; otherwise, it returns true.
For other directions, the corresponding illustrations can be obtained by rotating scenarios (a)-(p) from Figure~\ref{fig: checksplit3d} in the 3D space.

\section{Experiments}
We outline our experimental settings in Section~\ref{subsec: experiment settings}. Then, a comparison between 3D SEEDS and the SLIC algorithm~\cite{6205760} is presented, with quantitative results detailed in Section~\ref{subsec: quantitative results} and qualitative results detailed in Section~\ref{subsec: qualitative results}. 

\subsection{Experiment Settings}
\label{subsec: experiment settings}
\noindent\textbf{Metrics.}
We evaluate over-segmentation performance using the under-segmentation error proposed by SEEDS~\cite{van2015seeds} and introduce a new metric: the achievable Dice score. 
Denoting the supervoxel result as $\{A_{k}\}^{K}_{k=1}$, the ground-truth mask as $\{L_{i}\}^{I}_{i=1}$. each supervoxel can be aligned to the $i$-th ground-truth mask $L{i}$ with the largest overlap:
\begin{equation}
    i = \argmax_{i}|{A_{k} \cap L_{i}}|
\label{equ: superpixel to ground truth}
\end{equation}
$|\cdot|$ denotes the size of an area. The largest overlap $\mathcal{L}(A_{k})$ is computed as:
\begin{equation}
    \mathcal{L}(A_{k}) = \max_{i}|{A_{k} \cap L_{i}}|
\label{equ: largest overlap}
\end{equation}
The under-segmentation error~(UE) is computed as:
\begin{equation}
    UE = \frac{\sum_{k}{|A_{k} - \mathcal{L}(A_{k})|}}{\sum_{i}{|L_{i}|}}
\label{equ: under-segmentation error}
\end{equation}
Given Equation~\ref{equ: superpixel to ground truth}, a new supervoxel result $\{A_{i}\}^{I}_{i}$ can be obtained, which is the achievable over-segmentation performance. The achievable Dice score~(ADS) is computed as:
\begin{equation}
    ADS = \sum_{i} {\frac{2 \times |A_{i} \cap L_{i}|}{|A_{i}| + |L_{i}|}}
\label{equ: achievable dice score}
\end{equation}

\noindent\textbf{Datasets.}
We evaluate 3D SEEDS on 13 segmentation tasks across 10 organs.
For brain tumor segmentation, we use the BraTS21-Glioma dataset~\cite{baid2021rsna}, recently updated in the BraTS 2023 challenge\footnote[4]{\url{https://www.synapse.org/Synapse:syn51156910/wiki/622351}}, referred to as BraTS23-Glioma. 
The BraTS23-Glioma dataset includes data from 1251 patients. Each patient dataset contains multi-parametric MRI scans, including native (T1) and post-contrast T1-weighted (T1Gd), T2-weighted (T2), and T2 Fluid Attenuated Inversion Recovery (T2-FLAIR) volumes. 
There are 4 segmentation tasks, including the whole tumor (WT), GD-enhancing tumor (ET), peritumoral edematous/invaded tissue (ED), and necrotic tumor core (NCR).

For other segmentation tasks, we utilize the BTCV dataset\footnotemark[2], which contains CT scans from 30 patients. 
Each scan includes 13 abdominal organs and we evaluate 9 of them: spleen, right kidney, left kidney, gallbladder, liver, stomach, aorta, inferior vena cava, and pancreas.

\noindent\textbf{Implementation Details.}
For the BraTS23-Glioma dataset~\cite{baid2021rsna}, we clipped intensity values to the range between the 0.5-th and 99.5-th percentiles before rescaling them to the range [0, 1].
Note that, while technically feasible, we do not stack multiple modalities to form a multi-channel image volume.
For the BTCV dataset\footnotemark[2], we applied a window width of 400 and a window level of 40, followed by rescaling the intensity values to the range [0, 1]. 
For both datasets, the data was reshaped to $160 \times 160 \times 160$ using bicubic interpolation in-plane and nearest-neighbor interpolation along the third axis. 

During the experiment, the compactness parameter is set as 0.05 for the SLIC algorithm~\cite{6205760}. 
For 3D SEEDS, the boundary term weight $\lambda$ in Equation~\ref{equ: energy function} is set to 2, with 15 color histogram bins, 2 block-level updates, and a default of 4 pixel-level updates. 
Both algorithms are evaluated with 1000 and 4096 supervoxels. To align our evaluation within the deep learning framework, supervoxel generation is performed within a single worker of the PyTorch data loading framework.

\begin{table*}[!t]
\noindent
\begin{minipage}[t]{0.35\textwidth}
    \noindent
    \begin{minipage}[t]{1\textwidth}
        \centering
        \vspace{2pt}
        \caption{Results of supervoxel latency~(s).}
        \vspace{-7pt}
        \label{tab: latency}
        \setlength{\tabcolsep}{2pt}
        \small
        \newlength{\slic}
        \newlength{\seeds}
        \newlength{\iter}
        \newlength{\num}
        \settowidth{\slic}{SLIC}
        \settowidth{\seeds}{3D SEEDS}
        \settowidth{\iter}{+8 iters}
        \settowidth{\iter}{0.00}
        \begin{tabular}{c | c | c c c}
            \toprule
            \# & SLIC~\cite{6205760} & 3D SEEDS & +8 iters & +16 iters \\
            \midrule
            1000 & 8.22 & 0.62 & 1.86 & 3.06 \\
            4096 & 9.48 & 0.79 & 2.51 & 4.26 \\
            \bottomrule
        \end{tabular}
    \end{minipage}%
    \\
    \begin{minipage}[t]{1\textwidth}
        \centering
        \caption{Results of under-segmentation error~(\%). The best result is highlighted in \textbf{bold}, while the fastest 3D SEEDS configuration outperforming SLIC~\cite{6205760} is \underline{underlined}.}
        \vspace{-7pt}
        \label{tab: ue}
        \setlength{\tabcolsep}{2pt}
        \small
        \begin{tabular}{c | c | c c c}
            \toprule
            \# & SLIC~\cite{6205760} & 3D SEEDS & +8 iters & +16 iters \\
            \midrule
            1000 & 1.30 & \underline{1.18} & 1.07 & \textbf{1.04} \\
            4096 & 0.91 & \underline{0.75} & 0.69 & \textbf{0.67} \\
            \bottomrule
        \end{tabular}
    \end{minipage}%
\end{minipage}%
\hfill
\begin{minipage}[t]{0.31\textwidth}
    \centering
    \caption{Results of brain tumor segmentation with 1000 supervoxels. We show the achievable Dice score~(\%). The best result is highlighted in \textbf{bold}, while the fastest 3D SEEDS configuration outperforming SLIC~\cite{6205760} is \underline{underlined}.}
    \label{tab: brats 1000}
    \setlength{\tabcolsep}{3pt}
    \small
    \begin{tabular}{c c c c c}
        \toprule
        Method & WT & ET & ED & NCR \\
        \midrule
        SLIC~\cite{6205760} & 84.86 & 60.30 & 63.37 & 37.84 \\
        \midrule
        \vspace{2pt}
        3D SEEDS & 83.82 & 57.26 & \underline{63.38} & \underline{38.25} \\
        \vspace{3pt}
        +8 iters & \underline{84.99} & \underline{60.53} & 65.02 & 41.64 \\
        +16 iters & \textbf{85.25} & \textbf{61.62} & \textbf{65.45} & \textbf{42.95} \\
        \bottomrule
    \end{tabular}
\end{minipage}%
\hfill
\begin{minipage}[t]{0.31\textwidth}
    \centering
    \caption{Results of brain tumor segmentation with 4096 supervoxels. We show the achievable Dice score~(\%). The best result is highlighted in \textbf{bold}, while the fastest 3D SEEDS configuration outperforming SLIC~\cite{6205760} is \underline{underlined}.}
    \label{tab: brats 4096}
    \setlength{\tabcolsep}{3pt}
    \small
    \begin{tabular}{c c c c c}
        \toprule
        Method & WT & ET & ED & NCR \\
        \midrule
        SLIC~\cite{6205760} & 88.02 & 71.96 & 72.16 & 56.46 \\
        \midrule
        \vspace{2pt}
        3D SEEDS & \underline{88.44} & \underline{72.42} & \underline{73.46} & \underline{57.89} \\
        \vspace{3pt}
        +8 iters & 88.90 & 74.33 & 74.15 & 60.30 \\
        +16 iters & \textbf{89.02} & \textbf{74.68} & \textbf{74.34} & \textbf{61.29} \\
        \bottomrule
    \end{tabular}
\end{minipage}%
\end{table*}

\begin{figure}[!t]
\centering
    \includegraphics[width=0.96\columnwidth]{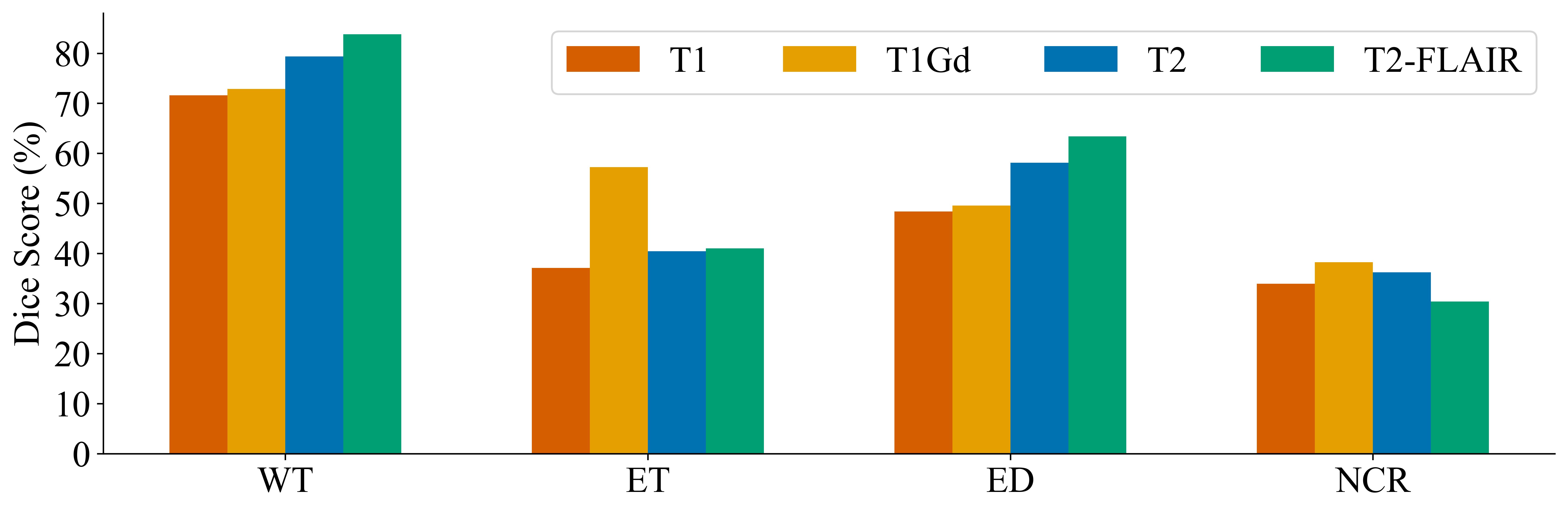}
    \caption{\textbf{Illustration of MRI Modalities \textit{v.s.} Brain Tumor Subregion Over-Segmentation Performance.} The evaluation uses the default 3D SEEDS configuration with 1000 supervoxels.}
    \label{fig: modalities}
\end{figure}

\begin{table*}[!t]
    \centering
    \caption{Results of abdominal organs segmentation with 1000 supervoxels. We show the achievable dice score~(\%). The best result is highlighted in \textbf{bold}, while the fastest 3D SEEDS configuration outperforming SLIC~\cite{6205760} is \underline{underlined}.}
    \label{tab: btcv 1000}
    \setlength{\tabcolsep}{4pt}
    \begin{tabular}{cccccccccc}
        \toprule
        Method & Spleen & Right Kidney & Left Kidney & Gallbladder & Liver & Stomach & Aorta & Inferior Vena Cava & Pancreas \\
        \midrule
        SLIC~\cite{6205760} & 69.91 & 56.09 & 46.29 & 5.24  & 81.54 & 50.51 & 5.24  & 1.15  & 0 \\
        \midrule
        3D SEEDS & \underline{75.96} & \underline{59.45} & \underline{60.39} & \underline{10.71} & \underline{84.76} & \underline{63.79} & \underline{17.10} & \underline{8.50}  & \underline{7.79}  \\
        +8 iters & 79.56 & 65.65 & 66.65 & 11.14 & 86.77 & 66.17 & 18.37 & \textbf{10.82} & \textbf{10.22} \\
        +16 iters & \textbf{80.54} & \textbf{66.58} & \textbf{67.99} & \textbf{11.47} & \textbf{87.41} & \textbf{66.66} & \textbf{20.13} & 9.39  & 10.14 \\
        \bottomrule
    \end{tabular}
\end{table*}

\begin{table*}[!t]
    \centering
    \caption{Results of abdominal organs segmentation with 4096 supervoxels. We show the achievable dice score~(\%). The best result is highlighted in \textbf{bold}, while the fastest 3D SEEDS configuration outperforming SLIC~\cite{6205760} is \underline{underlined}.}
    \label{tab: btcv 4096}
    \setlength{\tabcolsep}{4pt}
    \begin{tabular}{cccccccccc}
        \toprule
        Method & Spleen & Right Kidney & Left Kidney & Gallbladder & Liver & Stomach & Aorta & Inferior Vena Cava & Pancreas \\
        \midrule
        SLIC~\cite{6205760} & 81.60 & 79.02 & 80.24 & 30.89 & 87.88 & 73.25 & 51.32 & 31.59 & 22.76 \\
        \midrule
        3D SEEDS & \underline{86.67} & \underline{79.75} & 79.89 & \underline{43.10} & \underline{90.68} & \underline{79.06} & \underline{62.49} & \underline{50.72} & \underline{47.06} \\
        +8 iters & 88.54 & 82.35 & \underline{82.27} & 46.50 & 91.82 & 80.60 & 66.22 & 55.12 & 50.54 \\
        +16 iters & \textbf{88.94} & \textbf{82.94} & \textbf{82.77} & \textbf{46.81} & \textbf{92.07} & \textbf{80.95} & \textbf{67.12} & \textbf{56.31} & \textbf{52.17} \\
        \bottomrule
    \end{tabular}
\end{table*}

\subsection{Quantitative Results}
\label{subsec: quantitative results}

\noindent\textbf{Latency.} In Table~\ref{tab: latency}, we present the supervoxel generation latency. The latency is recorded within a single worker of the PyTorch data loading framework. Compared to SLIC~\cite{6205760}, the default 3D SEEDS configuration accelerates this process by a factor of 10. 3D SEEDS remains twice as fast as SLIC even with 16 additional pixel-level updates, which further improve the over-segmentation performance.

\noindent\textbf{UE.} In Table~\ref{tab: ue}, we present the under-segmentation error across all segmentation tasks. With 4096 supervoxels, 3D SEEDS with the default configuration reduces the under-segmentation error by 0.16\% when compared with SLIC~\cite{6205760}. This reduction can be further improved to 0.24\% with 16 additional pixel-level updates.

\noindent\textbf{ADS~(BraTS23-Glioma).}
In standard clinical protocol\footnotemark[3], ET and NCR are typically described by the T1Gd modality, while WT and ED are described by the T2-FLAIR modality. Figure~\ref{fig: modalities} supports this prior knowledge, demonstrating better performance on ET and NCR segmentation with the T1Gd modality and on WT and ED segmentation with the T2-FLAIR modality. The remaining experiments adhere to this rule.

In Tables~\ref{tab: brats 1000} and~\ref{tab: brats 4096}, we present the achievable Dice score on the BraTS23-Glioma dataset~\cite{baid2021rsna} with 1000 and 4096 supervoxels, respectively. 
The default 3D SEEDS configuration outperforms SLIC~\cite{6205760} baseline in most cases. Additional iterations of pixel-level updates consistently enhance performance, surpassing SLIC by an average of +2.46\% ADS.
We also observe that increasing the number of supervoxels significantly improves over-segmentation performance for both the SLIC and 3D SEEDS algorithms.
With more supervoxels, 3D SEEDS in its default configuration achieves a +12.37\% improvement in ADS, while SLIC achieves a +10.56\% improvement in ADS.
Notably, with 4096 supervoxels, 3D SEEDS achieves 89.02\% ADS on whole tumor segmentation, comparable to many deep learning-based methods~\cite{isensee2021nnu}, demonstrating its effectiveness for medical image analysis.


\noindent\textbf{ADS~(BTCV).}
In Table~\ref{tab: btcv 1000} and~\ref{tab: btcv 4096}, we present the achievable Dice score on the BTCV dataset\footnotemark[2] with 1000 and 4096 supervoxels, respectively. 
Equation~\ref{equ: superpixel to ground truth} does not align supervoxels to any ground-truth mask if it is significantly smaller than the supervoxel. Therefore, we only report results for organs that are sufficiently large under the current supervoxel setting. 
3D SEEDS consistently achieves superior over-segmentation performance on the BTCV dataset, demonstrating its generalizability to both MRI and CT scans. 
With the default configuration, 3D SEEDS outperforms SLIC~\cite{6205760} by an average of +8.52\% ADS. This improvement increases further to +11.99\% ADS with additional pixel-level updates.
Notably, 3D SEEDS achieves 88.94\% ADS on spleen segmentation and 92.07\% ADS on liver segmentation. 

\begin{figure*}[!t]
\centering
    \includegraphics[width=0.96\textwidth]{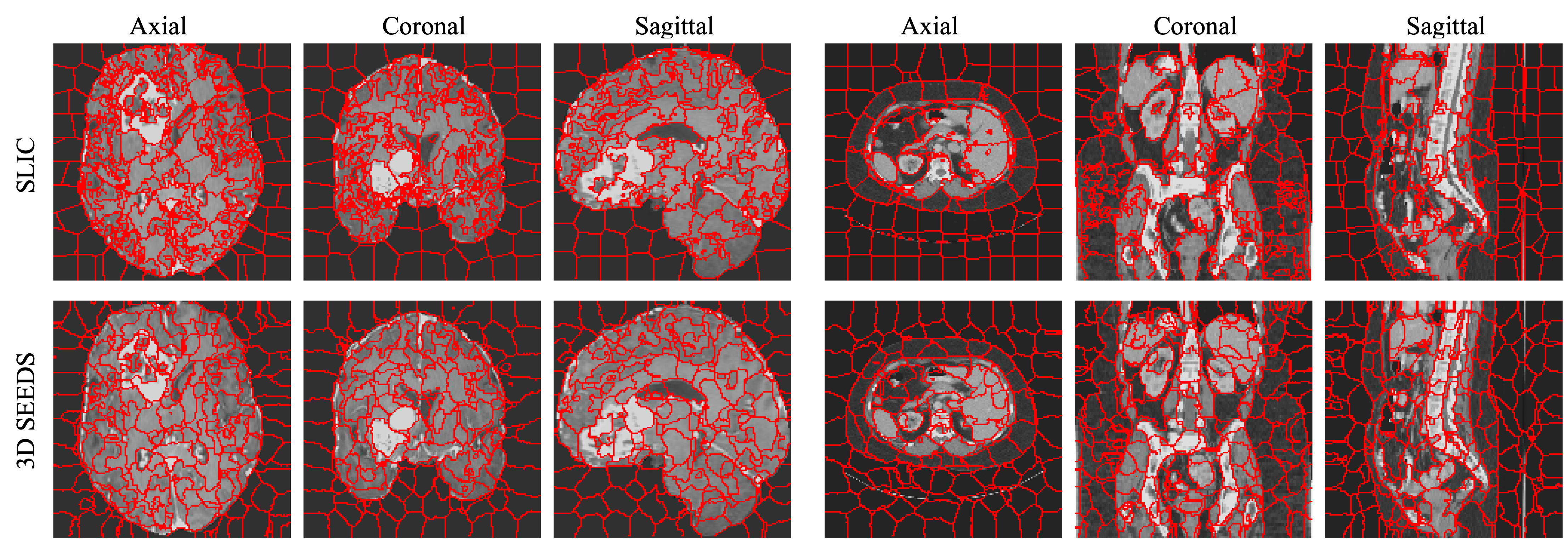}
    \caption{\textbf{Qualitative result with 1000 supervoxels.} Two examples each from the BraTS23-Glioma~(left) and BTCV~(right) datasets. The orientation for each example is set as ['R', 'P', 'I'], with visualizations including axial, coronal, and sagittal views.}
    \label{fig: vis 1000}
\end{figure*}
\begin{figure*}[!t]
\centering
    \includegraphics[width=0.96\textwidth]{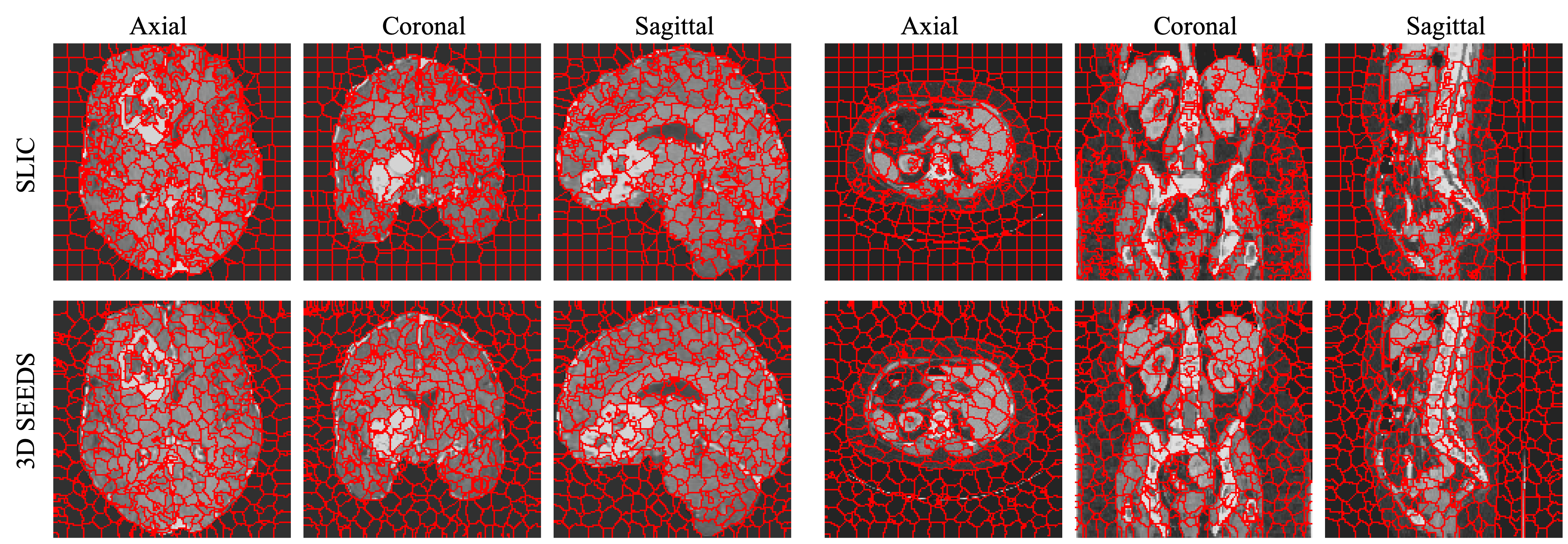}
    \caption{\textbf{Qualitative result with 4096 sueprvoxels.}Two examples each from the BraTS23-Glioma~(left) and BTCV~(right) datasets. The orientation for each example is set as ['R', 'P', 'I'], with visualizations including axial, coronal, and sagittal views.}
    \label{fig: vis 4096}
\end{figure*}

\subsection{Qualitative Results}
\label{subsec: qualitative results}

Figures~\ref{fig: vis 1000} and~\ref{fig: vis 4096} present the supervoxel results on the BraTS23-Glioma~\cite{baid2021rsna} and BTCV\footnotemark[2] datasets. To illustrate the results in 3D space, we include axial, coronal, and sagittal views. Compared to SLIC~\cite{6205760}, 3D SEEDS generates higher-quality supervoxels, especially when the number of supervoxels is relatively small~(\textit{i.e.}, 1000) or on the BTCV dataset. 
During the experiment, we observe that the number of block-level updates affects the qualitative performance when the number of supervoxels is set to 1000. This suggests that carefully tuning this parameter could further enhance the quantitative performance.

\section{Conclusion}

In this work, we bridge the gap between 2D superpixels and 3D supervoxels in the SEEDS algorithm, introducing the 3D SEEDS.
3D SEEDS is implemented closely adhering to the SEEDS algorithm provided by OpenCV and is compatible with both C++ and Python programming languages.
3D SEEDS achieves superior over-segmentation performance compared to the widely used SLIC supervoxel algorithm, even with the default configuration that accelerates supervoxel generation by a factor of 10.
In the context of deep learning, we perform 3D SEEDS within a single worker of the PyTorch data loading framework and achieve a supervoxel generation speed comparable to the NIfTI file read-in speed.
With additional update iterations, 3D SEEDS further improves over-segmentation performance while remaining twice as fast as the SLIC supervoxel algorithm.
3D SEEDS offers a flexible trade-off for deep-learning-based approaches: When speed is the priority, the default configuration enables online supervoxel generation. Conversely, when performance is more critical, supervoxels can be generated offline and yield better over-segmentation results.

{
    \small
    \bibliographystyle{ieeenat_fullname}
    \bibliography{main}
}


\end{document}